%% file: main.tex
\documentclass[10pt,twocolumn,letterpaper]{article}

\usepackage[pagenumbers]{iccv} 

\input{preamble}

\definecolor{iccvblue}{rgb}{0.21,0.49,0.74}
\usepackage[pagebackref,breaklinks,colorlinks,allcolors=iccvblue]{hyperref}

\def\paperID{14545} 
\def\confName{ICCV}
\def\confYear{2025}

\title{Is Extending Modality The Right Path Towards Omni-Modality?}

\author{
    Tinghui Zhu*\textsuperscript{\rm $\spadesuit$} \quad
    Kai Zhang*\textsuperscript{\rm $\clubsuit$} \quad
    Muhao Chen\textsuperscript{\rm $\spadesuit$} \quad
    Yu Su\textsuperscript{\rm $\clubsuit$} \\
\textsuperscript{\rm $\spadesuit$}{\small University of California, Davis} \quad
\textsuperscript{\rm $\clubsuit$}{\small The Ohio State University} \\
\;{\small \texttt{\{thuzhu, muhchen\}@ucdavis.edu}} \quad
{\small \texttt{\{zhang.13253, su.809\}@osu.edu}} 
}

\begin{document}
\maketitle
\input{sec/0_abstract}

\input{sec/1_1_intro}
\input{sec/2_relatedwork}

\input{sec/3_preliminary}
\input{sec/4_modality_finetuning}
\input{sec/5_model_merge}
\input{sec/6_pretrain}
\input{sec/7_conclusions}

\section*{Acknowledgements}
We would like to thank Xiang Yue (Carnegie Mellon University), Yue Bai (Northeastern University), and Chandan Singh (Microsoft Research) for their insightful discussions during the early stages of this work.
We also thank Zixiang Xu for his contributions during the initial exploration.
Finally, we thank the OSU NLP group for their constructive feedback and discussions.

\newpage
{
    \small
    \bibliographystyle{ieeenat_fullname}
    \bibliography{main}
}

\end{document}

%% file: preamble.tex
\usepackage{amsmath}

\usepackage{color}
\usepackage{colortbl}

\usepackage{booktabs}
\usepackage{multirow} 
\usepackage{soul}
\usepackage{xcolor} 
\usepackage{tabularx}

\definecolor{lightgreen}{RGB}{199,249,204}
\definecolor{lightblue}{RGB}{189,224,254}

\definecolor{color1}{HTML}{0a9396}
\definecolor{color2}{HTML}{94d2bd}
\definecolor{color3}{HTML}{e9d8a6}
\definecolor{color4}{HTML}{ee9b00}
\definecolor{color5}{HTML}{ca6702}
\definecolor{color6}{HTML}{bb3e03}

\newcolumntype{x}{>{\columncolor{color1!30}}c}
\newcolumntype{o}{>{\columncolor{color2!30}}c}
\newcolumntype{y}{>{\columncolor{color3!30}}c}
\newcolumntype{g}{>{\columncolor{color4!30}}c}
\newcolumntype{q}{>{\columncolor{color5!30}}c}
\newcolumntype{z}{>{\columncolor{color6!30}}c}

\newcommand{\SlightIncrease}{\textcolor{green!70!black}{$\uparrow$}}
\newcommand{\GreatIncrease}{\textcolor{green!50!black}{$\uparrow$}}
\newcommand{\SlightDecrease}{\textcolor{red!50}{$\downarrow$}}
\newcommand{\GreatDecrease}{\textcolor{red}{$\downarrow$}}
\newcommand{\Hold}{\textcolor{gray}{\footnotesize{=}}}

\usepackage{pifont}
\usepackage{bbm}

\usepackage{graphicx}

\usepackage{wrapfig}
\definecolor{caribbeangreen}{rgb}{0.0, 0.8, 0.6}

\usepackage{enumitem}
\usepackage{tcolorbox}
\usepackage{lipsum} 

\newcommand{\stitle}[1]{\vspace{0em} \noindent{\bf #1.}}

%% file: sec/0_abstract.tex
\begin{abstract}

Omni-modal language models (OLMs) aim to integrate and reason over diverse input modalities—such as text, images, video, and audio—while maintaining strong language capabilities. Despite recent advancements, existing models, especially open-source ones, remain far from true omni-modality, struggling to generalize beyond the specific modality pairs they are trained on or to achieve strong performance when processing multi-modal inputs.
We study the effect of extending modality, the dominant technique for training multimodal models, where an off-the-shelf language model is fine-tuned on target-domain and language data.
Specifically, we investigate three key questions: (1) Does modality extension compromise core language abilities? (2) Can model merging effectively integrate independently fine-tuned modality-specific models to achieve omni-modality? (3) Does omni-modality extension lead to better knowledge sharing and generalization compared to sequential extension?
Through extensive experiments, we analyze these trade-offs and provide insights into the feasibility of achieving true omni-modality using current approaches \footnote{Codes are avaliable at \url{https://github.com/DarthZhu/lm-extend}.}.
\end{abstract}

%% file: sec/1_1_intro.tex
\section{Introduction}
\label{sec:intro}

Omni-modal language models (OLMs) refer to models that can accept and understand various input modalities—text, images, video, audio, etc.—and engage with users with language in a seamless, natural manner.
Ideal OLMs are able to combine inputs from different modalities into a unified perception of real-world scenarios, enabling deeper contextual comprehension and more comprehensive reasoning.
This capability would empower embodied~\cite{ma2024survey} and virtual~\cite{deng2023mindweb} agents to perceive their environment.

\begin{figure}[t!]
    \centering
    \includegraphics[width=0.48\textwidth]{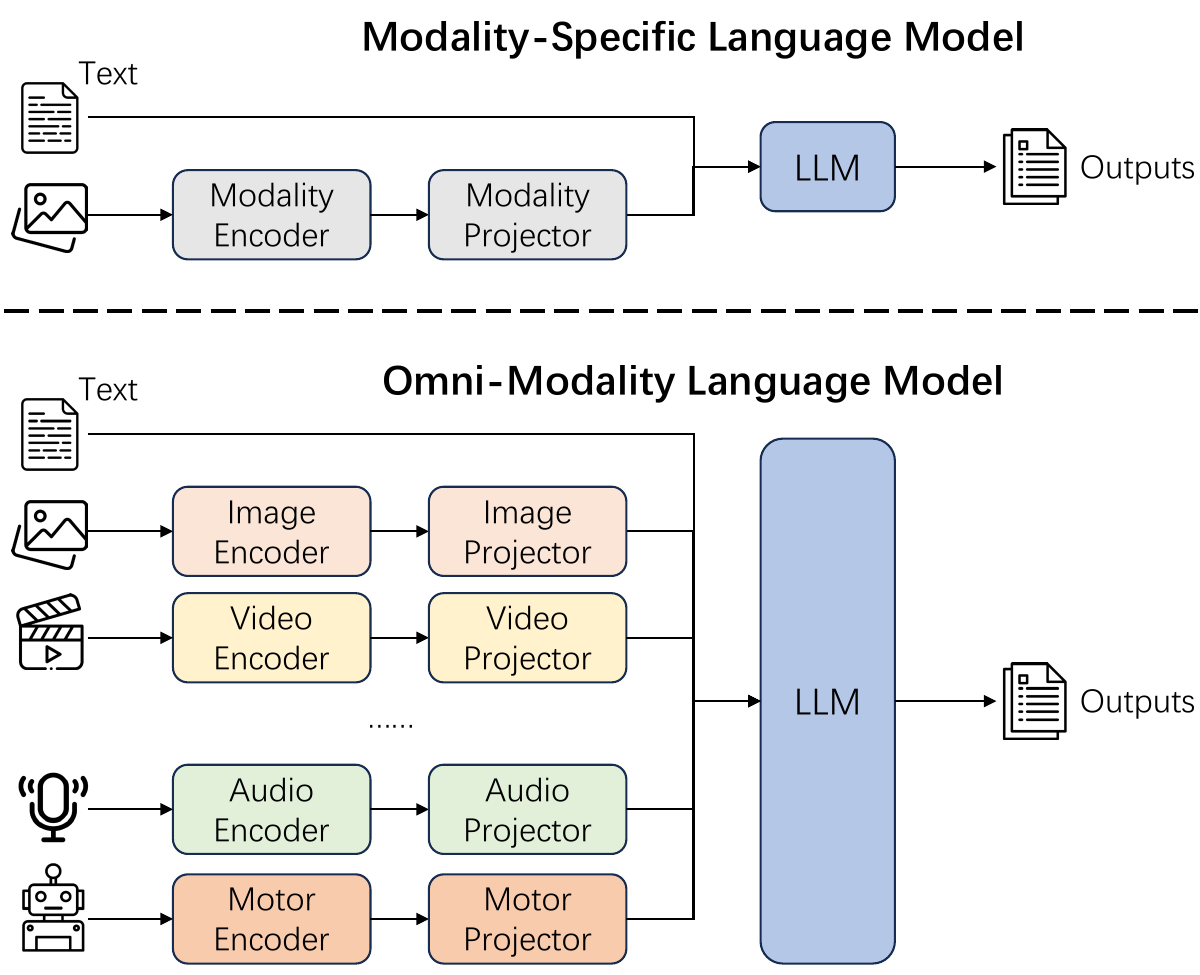}
    \caption{Overview of modality-specific language models and Omni-modal language models.}
    \label{fig:overview}
\end{figure}

OLMs belong to the broader category of multimodal models.
While recent advancements have made substantial progress in this field~\cite{openai2024gpt4v, li2024llava}, current models still lack true omni-modality—the ability to handle arbitrary modality combinations while maintaining robust reasoning and interaction abilities, evidenced by their inability to generalize beyond the specific modality pairs they were trained on.
For example, models trained on text-image tasks~\cite{liu2023visual, li2023blip} struggle with video understanding, and models optimized for text-video tasks~\cite{fu2024video, zhang2024video} often fail to incorporate spatial reasoning from static images.
Additionally, open-source OLMs~\cite{wu2024next} following often exhibit weaker performance on benchmarks specifically designed for fewer modality evaluation, such as text-image or text-video tasks, compared to models optimized for such tasks.

Notably, almost all existing multimodal models rely on a common strategy: \textit{extending modality}.
As shown in Figure~\ref{fig:overview}, this technique fine-tunes an off-the-shelf large language model (LLM) on data that pairs language with target modalities, enabling rapid adaptation to multimodal tasks~\cite{li2024llava, wu2024next}.
However, the extent to which extending modality contributes to the limitations of current multimodal models remains unclear (e.g., its impact on fundamental language capabilities).
Since LLMs serve as the backbone for most multimodal models, an important question is \textit{RQ1: whether modality extension compromises their core language abilities} in favor of a stronger performance on modality-specific benchmarks.
Do models retain their original reasoning and linguistic proficiency, or does the introduction of additional modalities interfere with their generalization across language tasks?

With the abundance of modality-extended models in the open-source community, another open question is: \textit{RQ2: Can we preserve their positive effects while extending multimodal capabilities using existing models?}
Model merging strategies~\cite{wortsman2022ModelSoups} have been explored in various applications to combine knowledge from different models, often improving performance on downstream tasks.
However, it remains unclear whether such techniques can be leveraged to create an effective OLM by merging multiple modality-extended models trained on different modalities.
Would merging independently fine-tuned models allow them to integrate cross-modal knowledge effectively, or would inconsistencies between separate modality-extended models hinder the fusion process?

Furthermore, \textit{RQ3: Does omni-modality fine-tuning lead to a more effective OLM?}
While some multimodal models have been trained to handle multiple modalities simultaneously, there has been no systematic comparison of different strategies for extending multiple modalities.
Most current approaches add one modality at a time through sequential fine-tuning~\cite{zhang2024video}, yet it remains unclear whether this stepwise process is more effective than omni-modality fine-tuning.
Would joint training across multiple modalities improve knowledge sharing and downstream performance, or is task-specific fine-tuning a more efficient approach?


Our findings highlight the impact of modality fine-tuning and the limitations it presents in the pursuit of OLMs.
First, we identify a trade-off between extending modalities and preserving the core language capabilities.
While modality fine-tuning can enhance certain LLM abilities, especially in areas like knowledge extension where visual modalities (e.g., images and videos) provide significant improvements, it tends to degrade crucial functions such as reasoning and instruction-following.
Second, we introduce and compare weighted average model merging with standard average merging.
Our results show that weighted model merging achieves the best performance across both textual and multimodal tasks, successfully preserving the most critical attributes of the original LLM, with parameter shifts acting as indicators of importance.
We also demonstrate that each attention head in modality fine-tuned models is integral to completing modality-specific tasks.
Third, we compare omni-modality fine-tuning with modality-specific fine-tuning, revealing that, while omni-modality fine-tuning holds conceptual appeal, it is less effective and efficient than models fine-tuned for specific modalities.
For instance, LLaVA-Next, fine-tuned for image tasks, outperforms NextGPT in image-based tasks using less data, highlighting the need for further investigation into omni-modality fine-tuning.
Moreover, we experiment the small-step fine-tuning \citep{cohere2025command} on the weighted merged model.
The results show that although small-step fine-tuning works on merged language models, it fails in omni-modal models.

In summary, this work examines the impact and potential of modality fine-tuning and omni-modality fine-tuning in extending LLM capabilities.
While modality fine-tuning improves specific abilities, it compromises others, and model merging, particularly weighted averaging, offers a promising strategy for maintaining multimodal capabilities.
However, merged models still fall short of the performance of modality-specific models, and omni-modality fine-tuning remains an area requiring more refined design and resource considerations.

%% file: sec/2_relatedwork.tex
\section{Related Work}
\label{sec:related}

\subsection{Multimodal Large Language Models}  
Multimodal large language models (MLLMs;~\cite{openai2024gpt4v,zhang2024mm}) have gained significant attention due to their ability to process and reason across different modalities, including text, images, video, and audio.
Recent approaches~\cite{alayrac2022flamingo,li2022blip} have explored fine-tuning language models with modality-specific data to enable them to handle multimodal inputs directly.
For instance, BLIP-2~\cite{li2023blip} and LLaVA~\cite{liu2023visual} focus on extending image understanding capabilities, while Video-LLaMA~\cite{zhang2023video} extends video understanding.
However, these models still suffer from modality issues, such as hallucinations~\cite{huang2025survey,huang2024visual} and knowledge conflicts~\cite{zhu2024unraveling}.
Furthermore, the impact of applying fine-tuning using modality-specific data on the original LLM remains unclear.

\subsection{Omni-Modal Language Models}  
Omni-modal language models (OLMs;~\cite{hurst2024gpt}) aim to create a unified framework that can simultaneously process and reason about various modalities without the need for separate specialized models for each type of input.
Recent studies, such as NextGPT~\cite{wu2024next} and OLA~\cite{liu2025ola}, have attempted fine-tuning with multiple modality-specific data, integrating text, image, and video understanding into a single LLM.
These models leverage shared latent spaces to enhance cross-modal understanding~\cite{lu2022unified}, enabling more coherent understanding across heterogeneous data.
However, the effectiveness and efficiency of extending multiple modalities from the base LLM remains unclear, leaving the effective training pattern of OLMs indefinitive.

\subsection{Continual Learning}  
Continual learning for large language models~\cite{wu2024continual,yang2025recent} focusing on enabling LLMs to learn from a continuous data stream over time, therefore enabling knowledge expansion~\cite{kowalczyk2024designing,wang2025bring} and conflict dissolving~\cite{du2024unlocking}.
Continual learning can be categorized into: continual pre-training~\cite{qin2023recyclable,yu2024information}, continual fine-tuning~\cite{wu2024llama,minixhoferzero}, and continual alignment~\cite{yao2023editing,zhang2024cppo}.
Recent study has revealed several drawbacks of continual learning, including performance deterioration on general benchmarks and weakened safety~\cite{li2025should}.
However, the impact of extending modality, which can be categorized as continual fine-tuning, is still under-explored.

%% file: sec/3_preliminary.tex
\section{Preliminary}
\label{sec:preliminary}

\subsection{Modality Extension}
Before discussing how MLLMs extend modalities based on LLMs, we first formalize the key components of MLLMs.

\smallskip
\stitle{MLLM Architecture}
Contemporary MLLMs~\cite{liu2023visual,li2023blip} adopt a general architecture which consists of a base LLM, modality encoders $\mathcal{E} = \{E_{m_1}, E_{m_2}, ..., E_{m_n}\}$, and modality projectors $\mathcal{P} = \{P_{m_1}, P_{m_2}, ..., P_{m_n}\}$, where each $m_i, i\in [1, 2, ..., n]$ is a modality except the textual one.
Given a multimodal input $\mathcal{X} = \{x_{m_0}, x_{m_1}, x_{m_2}, ..., x_{m_n}\}$, where $m_0$ is the textual modality, for each modality $m_i$, the MLLM first encode the modality input $x_{m_i}$ using $F$.
Then, a projector $P_{m_i}$ projects the encoded input to the textual modality as $t_{m_i}$.
In the meantime, the language tokenizer tokenizes the textual input $x_{m_0}$ into a token sequence $t_{m_0}$.
The LLM takes the encoded multimodal input $\mathcal{T} = \{t_{m_0}, t_{m_1}, t_{m_2}, ..., t_{m_n}\}$ and generates the output by the probability $p_\text{LLM}(y|\mathcal{T})$.

\smallskip
\stitle{Modality Fine-Tuning}
Modality fine-tuning is a common way to extend modalities on LLMs.
Specifically, modality fine-tuning leverages modality-specific instruction data---including various tasks on a certain modality---to fine-tune a LLM to capture the inputs encoded by modality encoders and modality projectors.
There are two popular ways to achieve this: to \emph{freeze} the LLM and only train the modality encoder and projector or to \emph{fine-tune} the LLM with modality instruction data.
Recent studies have pointed out that unfreezing the LLM for modality fine-tuning is essential for keeping the most desirable attributes of LLMs, such as in-context learning~\cite{lin2024vila}.
Thus, in this paper, we discuss modality fine-tuning in the context of unfreezing the LLM.

\smallskip
\stitle{Omni-Modality Fine-Tuning}
Omni-modality fine-tuning refers to a training paradigm that aims to integrate multiple modalities into a single model, enabling it to process and reason across diverse input types, such as text, image, audio, and video.
Unlike modality-specific fine-tuning which focuses on one modality at a time, omni-modality fine-tuning requires the model to learn representations that capture the relationships and interactions between different modalities simultaneously.
This approach attempts to create a unified framework capable of seamlessly understanding and responding to multimodal inputs without the need for separate specialized models.

\subsection{General Experimental Setup}
For all the experiments, we adopt the default generation parameters.
For multiple-choices datasets, we adopt greedy decoding to generate the options.
For datasets that require sampling, we set the temperature to 1.0.

\begin{table}[t!]
    \centering
    \input{src/table/4_datasets}
    \caption{Overview of datasets for evaluation on textual abilities.}
    \label{tab:4_datasets}
\end{table}

\begin{table*}[t!]
    \centering
    \input{src/table/4_models}
    \caption{Overview of base LLMs and their corresponding multimodal models (MLLMs), including the modalities they support and the associated modality extension data used for training.}
    \label{tab:4_models}
\end{table*}

%% file: src/table/4_datasets.tex
\begin{tabular}{lll}
\toprule
\textbf{Task}                 & \textbf{Dataset}    & \textbf{Size} \\ \midrule
\multirow{2}{*}{Knowledge}    & MMLU                & 14,079         \\
                              & MMLU-Pro            & 12,032         \\ \hline
Instruction Following         & IFEval              & 541           \\ \hline
\multirow{2}{*}{Long Context} & Passkey Retrieval   & 400           \\
                              & ZeroScrolls/Quality & 21            \\ \hline
\multirow{3}{*}{Reasoning}    & GPQA                & 198           \\
                              & MATH                & 5,000          \\
                              & HumanEval++         & 164           \\ \hline
Multilingual                  & MMMLU               & 196,588        \\ \hline
Safety                        & Harmbench           & 200           \\ \bottomrule
\end{tabular}

%% file: src/table/4_models.tex
\begin{tabular}{llll}
    \toprule
    \textbf{Base LLM}                  & \textbf{MLLM}               & \textbf{Modality} & \textbf{Modality Extension Data}            \\ \midrule
    \multirow{3}{*}{Qwen2-7B-Instruct} & Qwen2-VL-7B-Instruct        & Image \& Video    & \textgreater 1.4 trillion mutlimodal tokens \\
                                       & LLaVA-OneVision-Qwen2-7B-SI & Image             & 8.5m image data                             \\
                                       & LLaVA-Video-7B-Qwen2        & Video             & 8.5m image data + 1.6m image \& video data  \\
    \hline
    Qwen-7B-Instruct                   & Qwen2-Audio-7B-Instruct     & Audio             & 520k audio instruction pairs                \\
    \hline
    Vicuna-7B-V1.5                     & LLaVA-1.5-7B                & Image             & 600k image insturction pairs                \\
    \hline
    Qwen2-72B-Instruct                 & Qwen2-VL-72B-Instruct       & Image             & \textgreater 1.4 trillion multimodal tokens \\ 
    \bottomrule
\end{tabular}

%% file: sec/4_modality_finetuning.tex
\section{On the Impact of Modality Fine-Tuning on Base LLM}
\label{sec:modality_finetunig}
Supervised fine-tuning on specific modalities has proven effective in extending LLMs from purely textual to multimodal capabilities, especially with the LLM co-training with the modality encoder and projector.
However, modality fine-tuning without freezing the base LLM alters its default parameters, potentially affecting its original performance.
While some studies have discussed preserving the base LLM's capabilities, the broader implications of modality fine-tuning remain largely underexplored. 
In this section, we examine how fine-tuning on different modalities influences the base language model.
 
\subsection{Experimental Setup}

\stitle{Datasets}
To systematically assess the impact of modality fine-tuning, we evaluate six core LLM abilities: \colorbox{color1!30}{\emph{Knowledge}}, \colorbox{color2!30}{\emph{Instruction Following}}, \colorbox{color3!30}{\emph{Long Context}}, \colorbox{color4!30}{\emph{Reasoning}}, \colorbox{color5!30}{\emph{Multilingual}}, and \colorbox{color6!30}{\emph{Safety}}.  
\begin{itemize}
    \item For \colorbox{color1!30}{\emph{Knowledge}}, we adopt MMLU~\cite{hendrycks2020measuring} and MMLU-Pro~\cite{wang2024mmlu}.
    \item For \colorbox{color2!30}{\emph{Instruction Following}}, we adopt IFEval~\cite{zhou2023instruction}. We report an average performance across strict prompt, strict instruction, loose prompt, loose instruction.
    \item For \colorbox{color3!30}{\emph{Long Context}}, we adopt Passkey Retrieval~\cite{mohtashamilandmark} and the Quality subset of ZeroScrolls~\cite{shaham2023zeroscrolls}.
    \item For \colorbox{color4!30}{\emph{Reasoning}}, we evaluate three different domains, namely, general reasoning (GPQA~\cite{rein2024gpqa}), math reasoning (MATH~\cite{hendrycks2021measuring}), and coding (HumanEval-plus~\cite{liu2023your})
    \item For \colorbox{color5!30}{\emph{Multilingual}}, we adopt MMMLU~\cite{hendrycksmeasuring}.
    \item For \colorbox{color6!30}{\emph{Safety}}, we adopt HarmBench~\cite{mazeika2024harmbench}.
\end{itemize}
The statistics of these datasets are listed in \Cref{tab:4_datasets}.

\smallskip
\stitle{Models}
Supervised fine-tuning on different modalities can steer LLMs in diverse directions.
Intuitively, fine-tuning on image data may enrich the model’s contextual understanding, while video fine-tuning may enhance its ability to process long-range dependencies.
To systematically analyze these effects, we conduct controlled experiments across different modalities and model sizes.

Our primary analyses adopts the Qwen2-7B-Instruct model family~\cite{yang2024qwen2technicalreport}, as the following multimodal extensions of Qwen2-7B-Instruct easily supports a comprehensive comparison across modalities:
\begin{itemize}
    \item Image modality: Qwen2-VL-7B-Instruct~\cite{wang2024qwen2} and LLaVA-OneVision-Qwen2-7B-SI~\cite{li2024llava}.  
    \item Video modality: LLaVA-Video-7B-Qwen2~\cite{zhang2024video}.  
    \item Audio modality: Qwen2-Audio-7B-Instruct~\cite{chu2024qwen2}.  
\end{itemize}
To evaluate whether these findings generalize across different base LLMs, we also test Vicuna-7B-V1.5~\cite{vicuna2023}, alongside its image extension LLaVA-1.5-7B~\cite{liu2023visual}. 
Additionally, we assess the impact of model size by analyzing Qwen2-72B-Instruct~\cite{wang2024qwen2} and its visual extension, Qwen2-VL-72B-Instruct~\cite{yang2024qwen2technicalreport}.
\Cref{tab:4_models} shows the detailed model statistics.

\begin{table*}[ht!]
    \centering
    \small

\input{src/table/4_results}
    \caption{Performance of the base LLMs across all evaluated domains of textual abilities.}
    \label{tab:4_results}
\end{table*}

\subsection{Results}

The evaluation results are presented in \Cref{tab:4_results}, from which we derive several key observations.

\smallskip
\stitle{Visual modality extends the scope of parametric knowledge}
The results from the MMLU and MMLU-Pro datasets indicate that extending modality could effectively enhance the knowledge capabilities of LLMs, with improvements alongside different base language models and their model sizes.
Especially on the recent MMLU-Pro dataset, the visual (image and video) extensions of Qwen2 surpass the original Qwen2 by at least 2.5\%.  

A closer look at the modality-specific training data provides further insights.
The Qwen2-VL-7B-Instruct, which processes over 1.4T multimodal tokens, improves MMLU-Pro performance by approximately 5\%.
In contrast, the LLaVA-Video-7B-Qwen2, trained on around 10M instruction data (averagely less than 1.4T tokens), achieves a 2.5\% improvement.
This suggests that performance gains scale with the quantity of modality-specific training data, highlighting the hypothesis that visual modality fine-tuning effectively injects new knowledge that requires visual comprehension into the base LLM.

Conversely, the audio modality appears to contribute less to expanding the base model’s parametric knowledge.
On both MMLU and MMLU-Pro, Qwen2-Audio-7B-Instruct improves the average performance by only 0.4\% over its base model (Qwen2-7B-Chat).
This disparity suggests that, unlike the visual modality, which introduces novel form of knowledge (e.g., visual entity), the audio modality primarily serves as an extension of natural language, providing limited additional knowledge.

These findings imply that the ideal modality fine-tuning paradigms might be modality-specific.
For the audio modality, fine-tuning should prioritize alignment with the textual modality to enhance natural language understanding.
In contrast, visual modality fine-tuning should focus on synergizing multimodal representations to effectively integrate visual and textual knowledge.

\smallskip
\stitle{Modality fine-tuning harms instruction following, reasoning, and safety}
The results from the IFEval dataset reveal a significant performance decline in instruction following ability across all modality-extended models compared to the original base model.
Despite adopting an instruction-tuning format, modality fine-tuning fails to preserve the instruction-following capabilities of LLMs.
This suggests that current fine-tuning paradigms primarily function as modality extensions rather than serving a dual role in retaining instruction-following ability.
Consequently, incorporating instruction-following data during fine-tuning may be necessary to maintain this capability \citep{jindal2024balancing}.

Similarly, modality fine-tuning severely degrades reasoning performance across all tested domains, as evidenced by the 7B-scale results on GPQA, MATH, and HumanEval+.
The best-performing multimodal model, Qwen2-VL-7B-Instruct, still exhibits substantial performance drops: 3.0\% on GPQA, 10.2\% on MATH, and declines in both Pass@1 and Pass@5 on HumanEval+.
However, this degradation appears to diminish with model scaling.
On the Qwen2-VL-72B-Instruct model, the accuracy drop is reduced to 0.5\% for GPQA and 5.3\% for MATH, while the Pass@10 on HumanEval+ even surpasses that of the original Qwen2-72B-Instruct.
One possible explanation for this trend is that larger models possess more redundant parameters that are not salient to textual tasks.
These idle parameters may absorb the effects of modality fine-tuning, mitigating its impact on reasoning ability.
Nonetheless, despite this partial retention of the reasoning ability in larger models, the overall reasoning performance declines significantly after modality fine-tuning.

On the HarmBench dataset, the decline in alignment is evident across all evaluated models.
Except for the Qwen2-VL-7B-Instruct model, all multimodal models exhibit a significant increase in ASR@100, indicating weaker adherence to human moral alignment.
This observation aligns with findings from previous studies~\cite{lee2025how}, which suggest that modality fine-tuning disrupts the RLHF alignment, ultimately reducing the model’s safety compliance.

\stitle{Video modality may enhance the long context ability}
From the results of the ZeroScrolls dataset, we can observe that those models trained on large amounts of video data, except for those base LLMs do not have the long context ability, show an increase in the performance, \textit{i.e.} the Qwen2-VL-7B-Instruct and Qwen2-VL-72B-Instruct.
On the contrary, the LLaVA-OneVision model, which is trained on single image data, shows a decrease in the long context performance.
Intuitively, video is a visual version of the long-context document.
A typical video fine-tuning sample contains roughly 4k tokens \citep{zhang2024video}, mostly visual tokens.
Thus, training on video data could inherently enhance the long context understanding ability of the base LLM.

\stitle{Modality fine-tuning has a mixed effect on multilingual performance}
The results on the MMMLU dataset reveal varying effects of modality fine-tuning across different models and model sizes.
In vicuna-based models, the image extension enhances multilingual performance by 5.2\%, whereas in Qwen2-based models, performance either declines or remains unchanged.
However, in larger models, multilingual performance improves by 1.7\%.  

These findings suggest that the impact of modality fine-tuning on multilingual ability is not uniform and may depend on factors such as the base model architecture, pretraining data distribution, and the nature of the fine-tuning process.
The observed improvement in larger models could be attributed to their greater capacity to integrate multimodal information without compromising linguistic generalization.
Conversely, in smaller models, modality fine-tuning may introduce trade-offs that hinder multilingual performance, possibly due to parameter shifts that interfere with language-specific representations.



%% file: src/table/4_results.tex
\setlength{\tabcolsep}{1pt}
\resizebox{0.98\linewidth}{!}{
\begin{tabular}{lxxoyygggggqz}
\toprule
  \multirow{2}{*}{Model} &
  \cellcolor{white}MMLU &
  \cellcolor{white}MMLU-Pro &
  \cellcolor{white}IFEval &
  \cellcolor{white}PR &
  \cellcolor{white}ZeroScrolls &
  \cellcolor{white}GPQA &
  \cellcolor{white}MATH &
  \multicolumn{3}{c}{\cellcolor{white}HumanEval+} &
  \cellcolor{white}MMMLU &
  \cellcolor{white}HarmBench \\
\cmidrule(lr){2-2}\cmidrule(lr){3-3}\cmidrule(lr){4-4}\cmidrule(lr){5-5}\cmidrule(lr){6-6}\cmidrule(lr){7-7}\cmidrule(lr){8-8}\cmidrule(lr){9-11}\cmidrule(lr){12-12}\cmidrule(lr){13-13}
 &
  \cellcolor{white}Acc &
  \cellcolor{white}Acc &
  \cellcolor{white}Avg &
  \cellcolor{white}Acc &
  \cellcolor{white}Acc &
  \cellcolor{white}Acc &
  \cellcolor{white}Acc &
  \cellcolor{white}Pass@1 &
  \cellcolor{white}Pass@5 &
  \cellcolor{white}Pass@10 &
  \cellcolor{white}Acc &
  \cellcolor{white}ASR@100↓ \\
\hline
\multicolumn{13}{l}{\cellcolor{gray!30} \textbf{Qwen2-7B based}} \\
Qwen2-7B-Instruct &
  66.2 &
  35.1 &
  58.2 &
  100.0 &
  85.7 &
  13.6 &
  60.0 &
  54.2 &
  82.8 &
  93.4 &
  51.4 &
  15.9 \\
Qwen2-VL-7B-Instruct &
  66.5\SlightIncrease &
  40.0\GreatIncrease &
  48.6\GreatDecrease &
  100.0\Hold &
  90.5\GreatIncrease &
  10.6\GreatDecrease &
  49.8\GreatDecrease &
  37.6\GreatDecrease &
  75.4\GreatDecrease &
  92.1\SlightDecrease &
  51.4\Hold &
  11.4\GreatIncrease \\
LLaVA-Video-7B-Qwen2 &
  66.6\SlightIncrease &
  37.7\SlightIncrease &
  48.8\GreatDecrease &
  100.0\Hold &
  76.2\GreatDecrease &
  8.6\GreatDecrease &
  47.2\GreatDecrease &
  41.8\GreatDecrease &
  80.2\SlightDecrease &
  92.1\SlightDecrease &
  48.9\SlightDecrease &
  36.0\GreatDecrease \\
LLaVA-OneVision-Qwen2-7B-SI &
  65.4\SlightDecrease &
  37.9\SlightIncrease &
  28.6\GreatDecrease &
  100.0\Hold &
  76.2\GreatDecrease &
  4.6\GreatDecrease &
  14.6\GreatDecrease &
  0.0\GreatDecrease &
  0.0\GreatDecrease &
  0.0\GreatDecrease &
  48.2\SlightDecrease &
  34.1\GreatDecrease \\
\hline
\multicolumn{13}{l}{\textbf{\cellcolor{gray!30} Qwen-7B based}} \\
Qwen-7B-Chat &
  38.4 &
  14.9 &
  25.7 &
  35.5 &
  0.0 &
  5.1 &
  8.6 &
  0.0 &
  0.0 &
  0.0 &
  33.9 &
  8.2 \\
Qwen2-Audio-7B-Instruct &
  41.9\GreatIncrease &
  12.1\SlightDecrease &
  19.4\GreatDecrease &
  30.5\GreatDecrease &
  14.3\GreatIncrease &
  4.6\SlightDecrease &
  2.2\GreatDecrease &
  0.0\Hold &
  0.0\Hold &
  0.0\Hold &
  30.0\SlightDecrease &
  0.0\GreatIncrease \\
\hline
\multicolumn{13}{l}{\textbf{\cellcolor{gray!30} Vicuna-7B based}} \\
Vicuna-7B-V1.5 &
  45.5 &
  17.8 &
  41.8 &
  50.0 &
  0.0 &
  10.1 &
  6.1 &
  8.0 &
  24.3 &
  43.9 &
  30.1 &
  24.6 \\
LLaVA-1.5-7B &
  48.9\SlightIncrease &
  20.9\SlightIncrease &
  37.6\SlightDecrease &
  50.0\Hold &
  9.5\GreatIncrease &
  8.6\SlightDecrease &
  19.6\GreatIncrease &
  6.8\SlightDecrease &
  22.6\SlightDecrease &
  42.0\SlightDecrease &
  35.3\GreatIncrease &
  44.4\GreatDecrease \\
\hline
\multicolumn{13}{l}{\cellcolor{gray!30} \textbf{Qwen2-72B based}} \\
Qwen2-72B-Instruct &
  79.0 &
  48.7 &
  81.4 &
  100.0 &
  85.7 &
  10.1 &
  70.0 &
  72.5 &
  88.1 &
  93.4 &
  67.0 &
  1.5 \\
Qwen2-VL-72B-Instruct &
  81.5\SlightIncrease &
  50.2\SlightIncrease &
  62.9\GreatDecrease &
  100.0\Hold &
  90.5\GreatIncrease &
  9.6\SlightDecrease &
  64.7\GreatDecrease &
  47.7\GreatDecrease &
  88.0\SlightDecrease &
  96.7\SlightIncrease &
  68.7\SlightIncrease &
  11.2\GreatDecrease \\
\bottomrule
\end{tabular}
}

%% file: sec/5_model_merge.tex
\section{On Model Merging Towards an Omni-Modal Language Model}
\label{sec:model_merging}

\begin{figure*}[t]
    \centering
    \begin{subfigure}{0.33\textwidth}
        \centering
        \includegraphics[width=\linewidth]{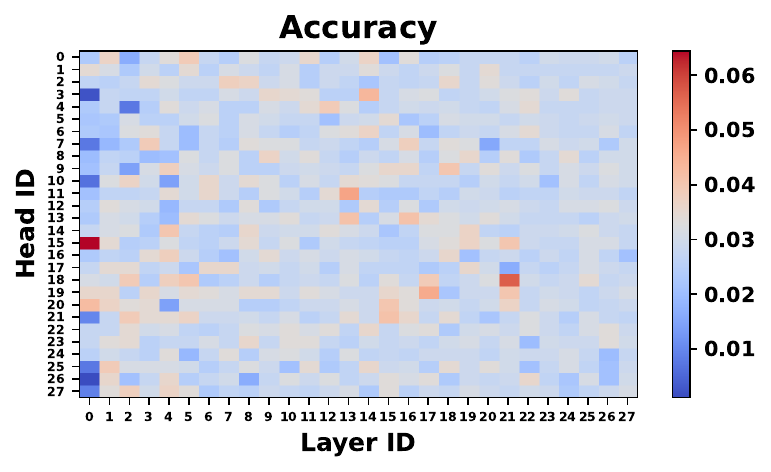}
    \end{subfigure}
    \hfill
    \begin{subfigure}{0.33\textwidth}
        \centering
        \includegraphics[width=\linewidth]{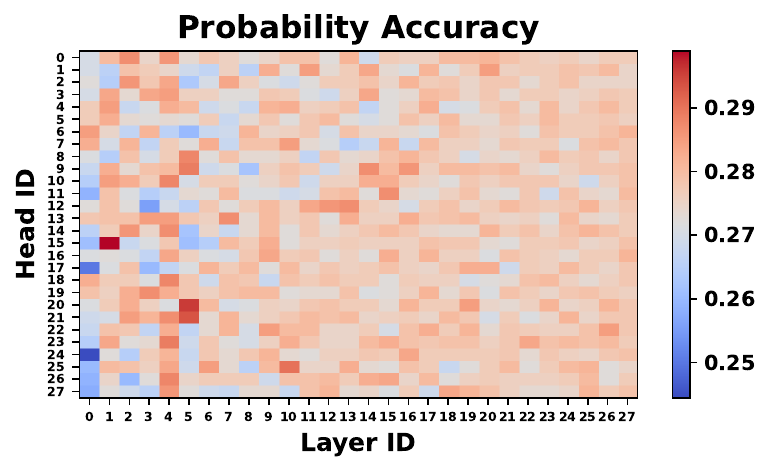}
    \end{subfigure}
    \hfill
    \begin{subfigure}{0.33\textwidth}
        \centering
        \includegraphics[width=\linewidth]{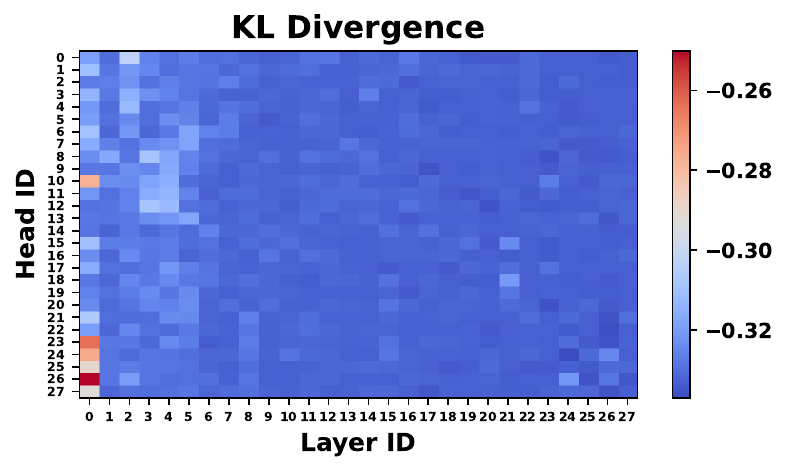}
    \end{subfigure}
    \caption{Heat map for masking each attention head. We report accuracy, accuracy calculated by probability, and KL divergence. The accuracy and probability accuracy should be as high as possible, while the KL divergence should have small absolute value.}
    \label{fig:head}
\end{figure*}

Having gained a clearer insight into the impact of modality fine-tuning, demonstrating both its benefits and drawbacks on the textual modality, we now explore a potential path towards omni-modal language models.
Specifically, we ask: \emph{Is it possible to preserve the positive effects and extend multimodal capabilities without further training the existing models?}  

A promising and lightweight approach to addressing this question is \emph{model merging}, which involves integrating the parameters of models with different training corpus and paradigms but the same architecture.
Model merging has been shown to be effective in various contexts, including knowledge editing~\cite{lu2025twin} and cross-modal knowledge transfer~\cite{ahmed2022cross}.
Thus, investigating model merging as a means of combining modality-specific expertise while mitigating the negative effects of fine-tuning presents an intriguing direction for developing more robust omni-modal models.
In the following sections, we explore different model merging strategies and evaluate their effectiveness in preserving textual capabilities while enabling multimodal capability.

\subsection{Merging Methods}  
We employ two widely used model merging techniques: \emph{average merging} and \emph{weighted average merging}, both of which are task- and modality-agnostic.  
\emph{Average merging} computes the element-wise average of the weights across all candidate models, while \emph{weighted average merging} assigns a heuristic weight to each model’s parameters before averaging.  
Formally, \emph{average merging} is defined as:  
\begin{equation}
    \theta_\text{merge} = \sum_{i=1}^n \theta_i,
\end{equation}
where $\theta_i$ represents the parameters of the $i^\text{th}$ candidate model.  
For \emph{weighted average merging}, the merged parameters are computed as:  
\begin{equation}
    \theta_\text{merge} = \sum_{i=1}^n \alpha_i \theta_i,
\end{equation}
where $\alpha_i$ denotes the weight assigned to the parameters of the $i^\text{th}$ model.  

To ensure the merging process remains both task- and modality-agnostic, the weight $\alpha_i$ is designed to be independent of specific tasks or modalities.
This guarantees that newly extended multimodal models can be seamlessly merged without adaptation to modality-specific tasks.

\subsection{Examination of Model Parameters}  
To determine the appropriate design for parameter weights in model merging, we must first answer a fundamental question: \emph{What is the largest unit to which model merging can be applied?}  
If modality fine-tuning affects only a subset of parameters, merging should ideally be constrained to these altered parameters while preserving the original ones.  

To investigate this, we analyze \emph{head-level modality salience}, which quantifies the contribution of individual attention heads to modality-specific tasks.
Specifically, we iteratively mask a single attention head and evaluate the resulting impact on model performance, allowing us to assess the relative importance of each head in processing multimodal information.
We conduct this analysis using the Qwen2-VL-7B-Instruct model on the MMMU dataset~\cite{yue2024mmmu}.  

We employ three metrics to examine head-level modality salience:  
\begin{itemize}
    \item \textbf{Accuracy.} The model is prompted to generate an answer choice directly, and accuracy is computed based on correct predictions. The baseline accuracy of Qwen2-VL-7B-Instruct on MMMU is 49.44\%.  
    \item \textbf{Probability Accuracy.} To mitigate the impact of potential degradation in generation quality caused by masking a single attention head, we analyze the logits of the first generated token, expected to correspond to the answer choice. Specifically, we extract the logits of the options (\textit{i.e.}, A, B, C, and D), apply softmax normalization, and compute accuracy.  
    \item \textbf{KL Divergence.} To quantify distributional shifts, we compute the Kullback-Leibler (KL) divergence between the option logits of the original model and those obtained after masking an attention head.  
\end{itemize}
The results are presented in \Cref{fig:head}, from which several key observations can be made. 

\begin{table*}[t!]
    \centering
    \small
    \input{src/table/5_textual_results}
    \caption{Performance of merged LLMs across all evaluated domains of textual abilities.}
    \label{tab:5_textual_results}
\end{table*}

Across all three evaluation metrics, masking any attention head results in a substantial performance drop, indicating that no single head is dispensable for specific modality processing, unlike retrieval or long context abilities~\cite{wu2025retrieval}.
This suggests that modality fine-tuning modifies the entire parameter set rather than only specific attention heads, implying that the model merging weight design should account for all parameters rather than a subset of them.

Additionally, a notable trend emerges: attention heads in shallower layers (closer to the input) exert a greater influence on multimodal performance compared to those in deeper layers.
This observation aligns with the established role of transformer layers, where shallow layers primarily focus on semantic understanding, while deeper layers perform integration and reasoning~\cite{Wang2024Grokking}.
Consequently, masking shallow-layer attention heads disrupts input processing, leading to failure in modality-specific tasks.
These findings underscore the importance of preserving early-layer representations when designing model merging strategies for multimodal extensions.

\subsection{Weighted Model Merging}  
Since attention heads are too coarse-grained for effective model merging, we refine our approach by considering parameter matrices.
To quantify the extent of parameter modifications due to modality fine-tuning, we compute $\Delta_\text{avg}$ for each tensor, defined as:  
\begin{equation}
    \Delta_\text{avg} = \text{avg} |\theta_\text{ori} - \theta_\text{mft}|,
    \label{eq:delta_avg}
\end{equation}  
where $\theta_\text{ori}$ represents the parameters of the original LLM, and $\theta_\text{mft}$ denotes those of the modality fine-tuned LLM.
This metric captures the average parameter shifts across the model after modality fine-tuning.  

Our analysis reveals that Qwen2-VL-7B-Instruct, which undergoes the most extensive modality fine-tuning, exhibits the largest parameter shift from its base LLM.
The average parameter shift of Qwen2-VL-7B-Instruct is 10 times larger than those of LLaVA-Video-7B-Qwen2 and LLaVA-OneVision-Qwen2-7B-SI.
This observation supports the hypothesis that greater specialization in a modality results in more substantial parameter deviations.  

Motivated by this insight, we incorporate $\Delta_\text{avg}$ into the weight design for model merging.
Specifically, for each model parameter $\theta_i$, we first compute $\Delta_\text{avg}^i$ using \Cref{eq:delta_avg}.
We then apply softmax to the set $\{\Delta_\text{avg}^1, \Delta_\text{avg}^2, ..., \Delta_\text{avg}^n\}$, transforming the values into a probability distribution $\{\alpha_1, \alpha_2, ..., \alpha_n\}$.  

To preserve the capabilities of the original LLM, we introduce a manually assigned weight $\alpha_0$ for its parameters. The remaining weights are rescaled by multiplying each $\alpha_i$ by $1-\alpha_0$, ensuring a controlled balance between the original and fine-tuned models. 
The final weighted-averaged parameter is thus formulated as:  
\begin{equation}
    \theta_\text{merge} = \alpha_0 \theta_0 + (1-\alpha_0)\sum_{i=1}^n \alpha_i \theta_i.
\end{equation}

\subsection{Experimental Setup}
We experiment model merging on the Qwen2-7B-Instruct based models, \textit{i.e.}, Qwen2-VL-7B-Instruct, LLaVA-Video-7B-Qwen2, and LLaVA-OneVision-Qwen2-7B-SI.
For the evaluation the original textual modality, we follow the setup from \Cref{sec:modality_finetunig}.
For the evaluation on other modalities, we adopt the image and video dataset, using MMMU~\cite{yue2024mmmu} and Video-MME~\cite{fu2024video}.
For generation configuration, we follow the same setup in \Cref{sec:preliminary}.

\begin{table}[t!]
    \centering
    \input{src/table/5_mm_results}
    \caption{Performance of OLMs based on merged LLMs across all evaluated domains of multimodal abilities.}
    \label{tab:5_mm_results}
\end{table}

\subsection{Results}  

The results of the textual evaluation are presented in \Cref{tab:5_textual_results}, while the multimodal evaluation results are shown in \Cref{tab:5_mm_results}. From these, we derive several key observations.  

\stitle{Model merging preserves the capabilities of base models}  
The textual evaluation results indicate that the merged model retains most of the original LLM’s capabilities, with improvements in certain domains. 

\begin{itemize}
    \item \textbf{Knowledge}: Modality fine-tuning has been shown to expand the model’s knowledge base. Notably, the merged model outperforms even the fine-tuned models, suggesting an enhanced integration of multimodal knowledge.  
    \item \textbf{Instruction Following}: While fine-tuned models exhibit a decline in instruction-following ability, merging with the original LLM not only restores but also slightly improves this capability.
    \item \textbf{Long Context}: The merged model maintains performance comparable to the original LLM, indicating that model merging does not degrade this ability.
    \item \textbf{Reasoning}: A consistent performance drop is observed across reasoning tasks following fine-tuning. However, model merging mitigates this decline to some extent. 
    \item \textbf{Multilingual}: Performance improves post-merging, suggesting that merging helps consolidate multilingual understanding. 
    \item \textbf{Safety}: The merged model preserves the alignment and safety characteristics of the original LLM.  
\end{itemize}

In summary, except for \emph{reasoning}, the merged model performs on par with or better than the original LLM across evaluated domains.
This suggests that future efforts to retain the base LLM’s capabilities should focus on addressing reasoning degradation during modality fine-tuning.

\stitle{Weighted model merging preserves more model abilities}
Both textual and multimodal evaluations demonstrate that weighted-average model merging achieves more robust performance.
This suggests that parameter shift serves as a crucial indicator of a parameter’s importance.
Furthermore, results indicate that merging a greater number of models further enhances overall performance, highlighting the potential of leveraging multiple modality-extended models to improve omni-modality.



%% file: src/table/5_textual_results.tex
\setlength{\tabcolsep}{1pt}
\begin{tabular}{lxxoyygggggqz}
\toprule
  \multirow{2}{*}{Model} &
  \cellcolor{white}MMLU &
  \cellcolor{white}MMLU-Pro &
  \cellcolor{white}IFEval &
  \cellcolor{white}PR &
  \cellcolor{white}ZeroScrolls &
  \cellcolor{white}GPQA &
  \cellcolor{white}MATH &
  \multicolumn{3}{c}{\cellcolor{white}HumanEval+} &
  \cellcolor{white}MMMLU &
  \cellcolor{white}HarmBench \\
\cmidrule(lr){2-2}\cmidrule(lr){3-3}\cmidrule(lr){4-4}\cmidrule(lr){5-5}\cmidrule(lr){6-6}\cmidrule(lr){7-7}\cmidrule(lr){8-8}\cmidrule(lr){9-11}\cmidrule(lr){12-12}\cmidrule(lr){13-13}
 &
  \cellcolor{white}Acc &
  \cellcolor{white}Acc &
  \cellcolor{white}Avg &
  \cellcolor{white}Acc &
  \cellcolor{white}Acc &
  \cellcolor{white}Acc &
  \cellcolor{white}Acc &
  \cellcolor{white}Pass@1 &
  \cellcolor{white}Pass@5 &
  \cellcolor{white}Pass@10 &
  \cellcolor{white}Acc &
  \cellcolor{white}ASR@100↓ \\
\hline
Qwen2-7B-Instruct &
  66.2 &
  35.1 &
  58.2 &
  100.0 &
  85.7 &
  13.6 &
  60.0 &
  54.2 &
  82.8 &
  93.4 &
  51.4 &
  15.9 \\ 
\hline
\multicolumn{13}{l}{\cellcolor{gray!30} \textbf{Average}} \\
Qwen2-Text/VL/Video &
  \textbf{68.7} &
  31.0 &
  58.0 &
  100.0 &
  \underline{81.0} &
  \underline{10.1} &
  55.9 &
  \underline{49.6} &
  \underline{83.2} &
  \textbf{94.5} &
  \underline{52.7} &
  \underline{18.2} \\
Qwen2-Text/VL/Video/SI &
  68.5 &
  \underline{\textbf{37.4}} &
  \textbf{58.2} &
  100.0 &
  \textbf{85.7} &
  \textbf{11.1} &
  \underline{56.1} &
  49.5 &
  \textbf{83.3} &
  \textbf{94.0} &
  52.6 &
  20.4 \\
Qwen2-VL/Video &
  \textbf{68.7} &
  36.0 &
  54.8 &
  100.0 &
  \textbf{85.7} &
  9.6 &
  52.9 &
  46.0 &
  82.2 &
  93.2 &
  52.5 &
  19.4 \\
Qwen2-VL/Vide/SI &
  68.4 &
  \textbf{40.0} &
  53.6 &
  100.0 &
  \textbf{85.7} &
  9.6 &
  53.9 &
  47.8 &
  82.9 &
  \underline{94.2} &
  52.3 &
  23.7 \\
\multicolumn{13}{l}{\cellcolor{gray!30} \textbf{Weighted Average}} \\
Qwen2-Text/VL/Video/SI &
  \underline{68.6} &
  36.3 &
  \underline{58.1} &
  100.0 &
  \underline{81.0} &
  9.1 &
  \textbf{57.0} &
  \textbf{50.0} &
  82.2 &
  93.8 &
  \textbf{52.8} &
  \textbf{15.4} \\ 
\bottomrule
\end{tabular}

%% file: src/table/5_mm_results.tex
\begin{tabular}{l|cc}
    \toprule
    \textbf{Model}            & \textbf{MMMU} & \textbf{Video-MME} \\ \midrule
    Qwen2-VL-7B-Instruct & 49.44         & 62.84              \\ 
    Qwen2-avg-all            & 48.78         & 56.89              \\
    Qwen2-weighted-all       & 48.11         & 61.04              \\ \bottomrule
\end{tabular}

%% file: sec/6_pretrain.tex
\section{On Omni-Modality Fine-Tuning towards Omni-Modal Model}
Previous sections indicate that model merging still has some degradation in performance.
Thus, our next question is: \emph{Is omni-modality fine-tuning the right path towards OLM?}
In this section, we will discuss about the effectiveness and the efficiency of omni-modality fine-tuning.

\subsection{Modality Fine-Tuning on Language Model}
\subsubsection{Experimental Setup}

\stitle{Model}
For the choice of omni-modality fine-tuned models, we adopt NextGPT~\cite{wu2024next} and Qwen2.5-Omni \citep{xu2025qwen2}.
The former utilizes a frozen language backbone, while the latter trains the whole model.
For the choice of modality fine-tuned models, we choose models that is specialized in certain modality and has the same base LLM as NextGPT, including InstructBlip~\cite{dai2023instructblipgeneralpurposevisionlanguagemodels}, LLaVA-Next~\cite{liu2024improved}, Video-LLaMA~\cite{zhang2023video}, Video-LLaVA~\cite{zhang2024video}, and Vista-LLaMA~\cite{ma2023vista}.

\stitle{Datasets}
For easier comparison, we adopt the datasets that are used to evaluate NextGPT.
For image modality, we adopt VizWiz~\cite{bigham2010vizwiz} and VQAv2~\cite{goyal2017making}.
For video modality, we adopt MSVD-QA and MSRVTT-QA~\cite{xu2017video}.

\begin{table}[h]
    \centering
    \input{src/table/6_pretrain}
    \caption{Performance of Omni-modal language models and modality-specific language models on image and video domains. \colorbox{color6!30}{Red} indicates omni-modal language models and \colorbox{color1!30}{blue} indicates modality-specific language models.}
    \label{tab:6_pretrain}
\end{table}

\begin{table*}[t!]
    \centering
    \small
    \input{src/table/6_omni_comparision}
    \caption{Performance of the merged omni-modal model compared to the fine-tuned omni-modal model on language benchmarks. Abs. stands for the absolute difference and Rel. stands for relative difference.}
    \label{tab:6_omni_comparision}
\end{table*}

\subsubsection{Results}
Our experimental results are presented in \Cref{tab:6_pretrain}, detailing the training data volumes and model performance across evaluation datasets, and \Cref{tab:6_omni_comparision}, which compares the language capabilities of our merged and fine-tuned omni-modal models.
The findings reveal that omni-modal fine-tuning is currently less effective and efficient than modality-specialized models.
Furthermore, both model merging and omni-modal fine-tuning tend to degrade the original language capabilities.

For image-based tasks, LLaVA-Next requires only one-third of the training data used by NextGPT yet significantly outperforms it on visual understanding benchmarks.
Similarly, for video-based tasks, Vista-LLaMA achieves comparable performance to NextGPT while consuming only half the training data.  
These results suggest that while omni-modality fine-tuning serves as a proof-of-concept for generalizing across modalities, it requires a more refined design to achieve efficiency and performance on par with specialized models.
Further research is needed to optimize omni-modality fine-tuning strategies, ensuring they can effectively balance generalization and efficiency without excessive data consumption.

For comparison on language tasks, both the merged model and the fine-tuned model demonstrate a decline across most language abilities, particularly in reasoning tasks.
However, the merged model shows a slight improvement in language understanding and knowledge-related capabilities.
The averaged performance drop for the fine-tuned model is -6.3\%, compared to -2.6\% for the merged model.
This indicates that while both methods impact language skills, the fine-tuning approach appears to be more detrimental to core language abilities than model merging.

\subsection{Modality Fine-Tuning on Merged Model}
Given that model merging alone is not consistently effective in extending language models to multiple modalities while simultaneously maintaining their core language proficiencies, we shift our focus to employing small-step fine-tuning on the merged model.
Previous work \citep{cohere2025command} has demonstrated that fine-tuning merged language models with a small number of training steps can enhance their performance across various language-centric abilities.
Consequently, we explore whether this conclusion still stands for the multimodal situation.

\subsubsection{Experimental Setup}

\stitle{Model}
For the base model to conduct fine-tuning on, we utilize the weighted-average merged model detailed in \Cref{sec:model_merging}, \textit{i.e.}, Qwen2-weighted-all.

\stitle{Fine-Tuning Dataset}
The fine-tuning dataset comprises selections for distinct modalities: MetaMath \citep{yu2023metamath} for language, VisualWebInstruct \citep{jia2025visualwebinstruct} for image, and LLaVA-Video-178K \citep{zhang2024video} for video.
These datasets are chosen because: 1) they are curated for tasks requiring complex reasoning within their specific modality, and 2) they offer a standardized fine-tuning format, facilitating reproducible research.
To approximate a balanced token exposure across modalities, we set the data proportion for fine-tuning as $\text{text} : \text{image} : \text{video} = 3 : 2 : 1$.

\subsubsection{Analysis}

\begin{figure}[h]
    \centering
    \includegraphics[width=.9\columnwidth]{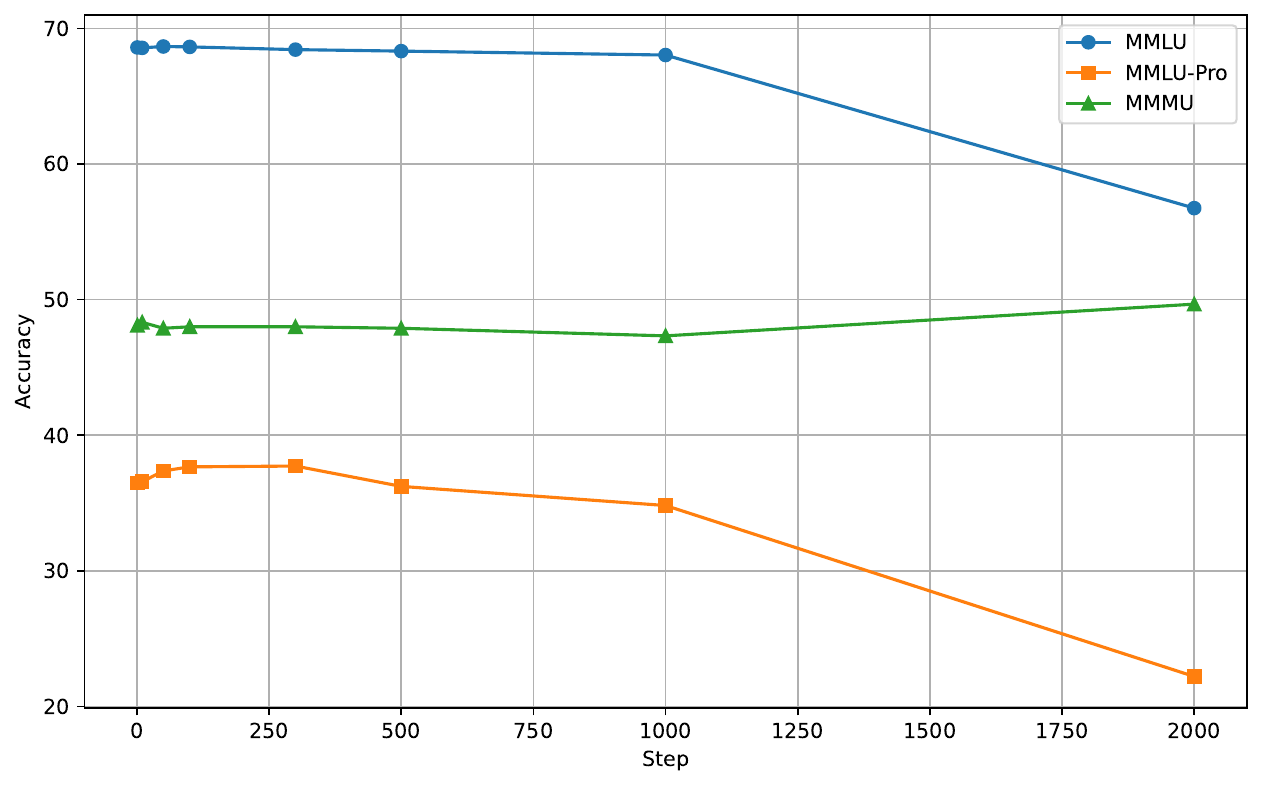}
    \caption{Model performance after n-step fine-tuning.}
    \label{fig:finetune}
\end{figure}

\stitle{Effects of Fine-Tuning Step Number}
To ground our analysis, we first seek to identify an optimal range for the number of fine-tuning steps.
We define optimal small-step fine-tuning as a regimen that 1) enhances modality alignment and performance on multimodal tasks, while 2) not substantially degrading the model's original language capabilities.
Performance is evaluated on the MMLU and MMLU-Pro benchmarks and the MMMU benchmark.

The results, depicted in \Cref{fig:finetune}, reveal a critical trend.
We observe that performance on both MMLU and MMLU-Pro tends to decrease after approximately 1,000 fine-tuning steps.
In contrast, performance on MMMU generally shows improvements with fine-tuning.
This divergence strongly suggests a \emph{modality trade-off}: enhancing multimodal understanding through fine-tuning can come at the cost of textual understanding.
This observation implies that straightforward small-step fine-tuning may present challenges for developing truly omni-modal models that excel universally across all modalities.
Furthermore, textual understanding (MMLU/MMLU-Pro) exhibits a slight increase or peak performance within the initial 100 steps, suggesting that the optimal fine-tuning step number for preserving or enhancing language abilities is relatively small.
Conversely, visual and multimodal capabilities (MMMU) may benefit from more fine-tuning steps.
This disparity in optimal fine-tuning step number for different modalities likely contributes to the observed trade-off.

\begin{figure}[htbp]
    \centering
    \begin{subfigure}[b]{0.48\columnwidth}
        \centering
        \includegraphics[width=\textwidth]{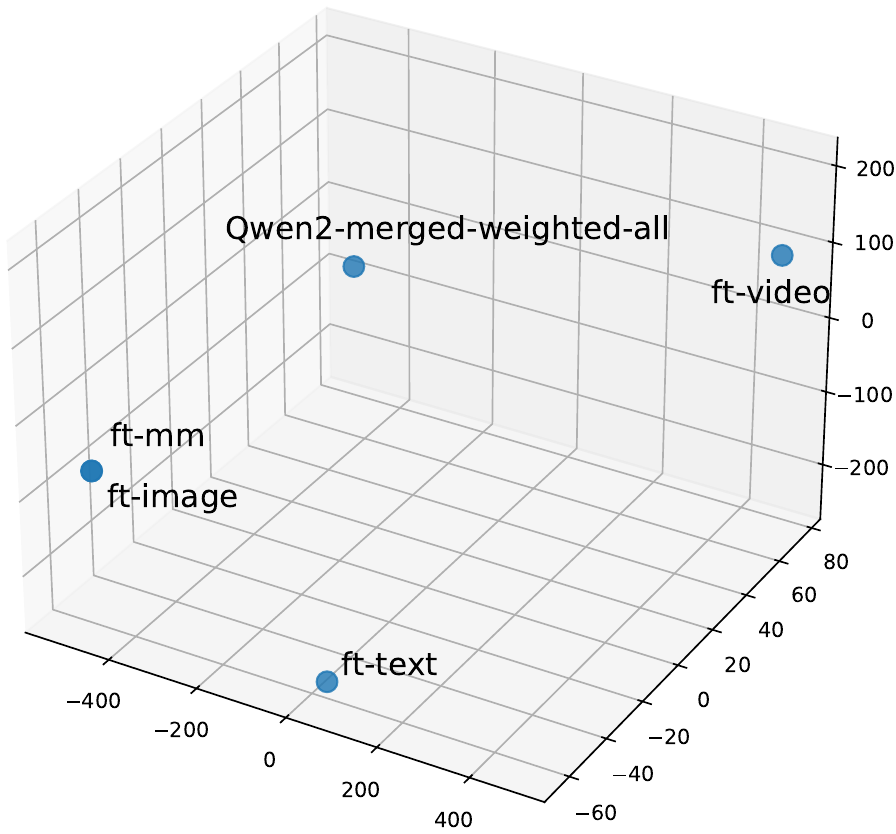}
        \caption{Visualization of weight shifts after fine-tuning on text, image, video, and mixed modality datasets.}
        \label{fig:tsne_finetune}
    \end{subfigure}
    \hfill
    \begin{subfigure}[b]{0.48\columnwidth}
        \centering
        \includegraphics[width=\textwidth]{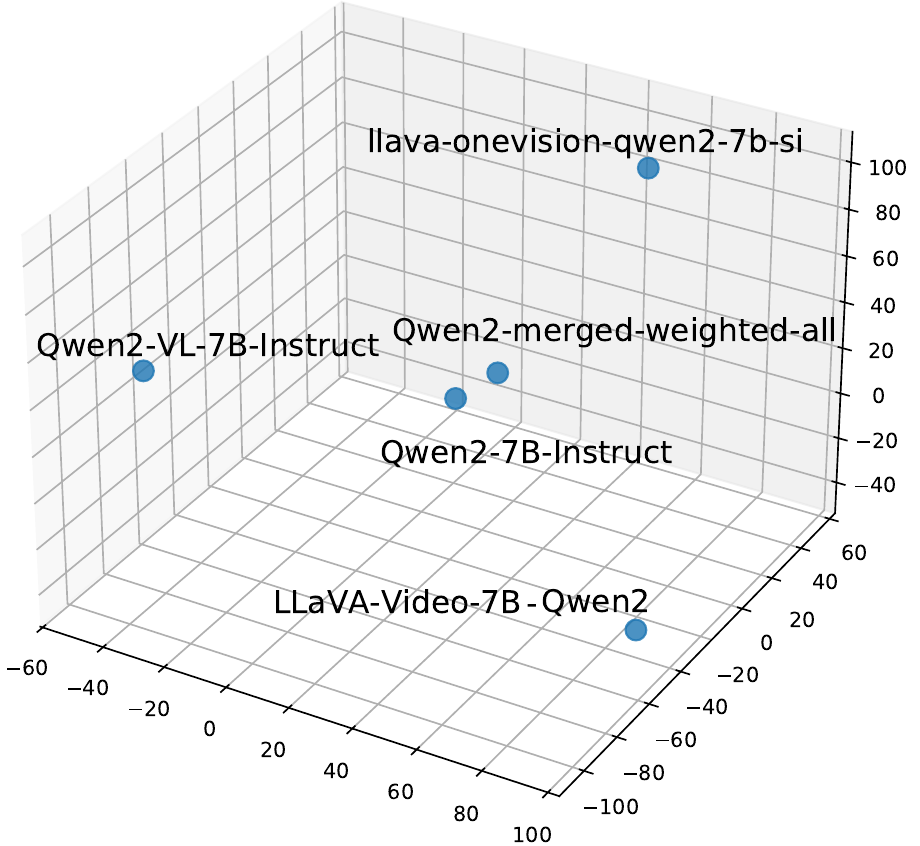}
        \caption{Visualization of relative positions of the base models and the weighted-average merged model.}
        \label{fig:tsne_merge}
    \end{subfigure}
    \caption{Comparative visualizations of model weight distributions. We use t-SNE to visualize the weight shifts.}
    \label{fig:weight-tsne}
\end{figure}

\stitle{Effects of Fine-Tuning V.S. Merging}
To further understand the mechanisms contributing to the \emph{modality trade-off}, we investigate the shifts in model weights induced by the merging process itself versus subsequent fine-tuning.
For the fine-tuning aspect of this experiment, we sample 1,000 instances from each modality-specific dataset (text, image, and video) and fine-tune the merged model (\textit{Qwen2-weighted-all}) on these individual sets, as well as on a combined mixed-modality set comprising all 3,000 samples.

The t-SNE \citep{van2008visualizing} visualizations of the weight distributions are presented in \Cref{fig:weight-tsne}.
From \Cref{fig:tsne_finetune}, it is evident that fine-tuning on different modalities propels the model weights in distinct directions within the parameter space.
This suggests that fine-tuning encourages specialization towards the statistical properties of the specific modality it is trained on.
Conversely, \Cref{fig:tsne_merge} indicates that weighted model merging positions the resultant model in a region that aggregates the weights of the base models.

This fundamental distinction in how weights are manipulated--fine-tuning driving towards specialized, often divergent points in the weight space versus merging seeking a consensual, interpolated representation--offers a compelling explanation for the observed modality trade-off.
While fine-tuning can significantly enhance performance for a target modality, it risks pulling the model's capabilities away from others.
Model merging, on the other hand, achieves an initial balance but may not unlock peak performance for any single modality.
Subsequent fine-tuning of this merged model, as shown, tends to quickly re-specialize the model, often reintroducing the trade-off by favoring improvement in one area at the expense of another.

%% file: src/table/6_pretrain.tex
\begin{tabular}{llcc}
    \toprule
    \multicolumn{4}{c}{\textbf{Text-Image}}         \\ \hline
     Model       & Data  & VizWiz  & VQAv2     \\ \hline
    \rowcolor{color6!30}
    NextGPT      & 4.5M  & 48.40   & 66.70     \\
    \rowcolor{color1!30}
    InstructBlip & 10M   & 34.50   & 43.30     \\
    \rowcolor{color1!30}
    LLaVA-Next   & 1.3M  & 57.60   & 81.80     \\ \midrule
    \multicolumn{4}{c}{\textbf{Text-Video}}         \\ \hline
     Model       & Data  & MSVD-QA & MSRVTT-QA \\ \hline
    \rowcolor{color6!30}
    NextGPT      & 2.1M  & 64.50   & 61.40     \\
    \rowcolor{color1!30}
    Video-LLaMA  & 2.8M & 51.60   & -         \\
    \rowcolor{color1!30}
    Video-LLaVA  & 2M    & 70.70   & 59.20     \\
    \rowcolor{color1!30}
    Vista-LLaMA  & 1.3M  & 65.30   & 60.50     \\ \bottomrule
\end{tabular}

%% file: src/table/6_omni_comparision.tex
\newcommand{\theadfont}{\normalsize\bfseries} 
\setlength{\tabcolsep}{4pt} 

\begin{tabular}{lccccccccc}
\toprule
 & \multicolumn{4}{c}{\textbf{Omni-Modality Fine-tuning}} & \multicolumn{4}{c}{\textbf{Weighted Merging}} \\
 \cmidrule(r){2-5} \cmidrule(l){6-9}
 & Qwen2.5-7B & Qwen2.5-Omni-7B & Abs. & Rel. & Qwen2-7B-Inst. & Qwen2-7B-weighted & Abs. & Rel. \\
\midrule
MMLU-Pro  & 56.3 & 47.0 & -9.3 & -16.5\% & 35.1 & 36.3 & +1.2 & +3.4\% \\
GPQA      & 36.4 & 30.8 & -5.6 & -15.4\% & 13.6 & 9.1  & -4.5 & -33.1\% \\
MATH      & 75.5 & 71.5 & -4.0 & -5.3\%  & 60.0 & 57.0 & -3.0 & -5.0\% \\
HumanEval & 84.8 & 78.7 & -6.1 & -7.2\%  & 54.2 & 50.0 & -4.2 & -7.7\% \\
\bottomrule
\end{tabular}

%% file: sec/7_conclusions.tex
\section{Conclusion}  

This work explore the impact of modality fine-tuning on LLMs and evaluated two alternative approaches for developing Omni-Modality Language Models (OLMs): model merging and omni-modality fine-tuning.
Modality fine-tuning effectively extends the capabilities of a base LLM to handle multimodal inputs but inevitably alters its parameters.
This modification can lead to both improvements in certain domains, such as knowledge expansion, and degradations in core abilities like reasoning and instruction following.
Weighted model merging mitigates some of these losses but does not fully preserve all capabilities.
Omni-modality fine-tuning, though conceptually promising, proves inefficient compared to modality-specialized models, requiring more training data while offering limited improvements.
Overall, our findings suggest that neither modality fine-tuning nor naive omni-modality fine-tuning offers a definitive solution to achieving robust OLMs.
We hope this study provides valuable insights for advancing research in multimodal LLMs and inspires new approaches toward achieving truly omni-modality models.